\documentclass[manuscript,screen]{acmart}

\usepackage{booktabs}   
\usepackage{wasysym}   
\usepackage{enumitem}

\AtBeginDocument{%
  }

\copyrightyear{2026}
\acmYear{2026}
\setcopyright{acmlicensed}
\acmDOI{https://doi.org/10.1145/3797246.3806223}
\acmConference[ETRA '26]{2026 Symposium on Eye Tracking Research and Applications}{June 01--04, 2026}{Marrakesh, Morocco}
\acmBooktitle{2026 Symposium on Eye Tracking Research and Applications (ETRA '26), June 01--04, 2026, Marrakesh, Morocco}
\acmISBN{979-8-4007-2519-7/2026/06}

\begin{document}

\title{What They Saw, Not Just Where They Looked: Semantic Scanpath Similarity via VLMs and NLP metrics}

\author{Mohamed Amine KERKOURI}
\orcid{1234-5678-9012}
\authornotemark[1]
\affiliation{%
  \institution{F-Initiatives}
  \city{Paris}
  \country{France}
}

\author{Marouane Tliba}
\affiliation{%
  \institution{USPN}
  \city{Paris}
  \country{France}}

\author{Bin Wang}
\affiliation{%
  \institution{Northwestern University}
  \city{Evanston, Illinois}
  \country{United States}
}

\author{Aladine Chetouani}
\affiliation{%
  \institution{USPN}
  \city{Paris}
  \country{France}
}

\author{Ulas Bagci}
\affiliation{%
  \institution{Radiology, Northwestern University}
  \city{Chicago, Illinois}
  \country{United States}
}

\author{Alessandro Bruno}
\affiliation{%
  \institution{IULM}
  \city{Milano}
  \country{Italy}
}

\renewcommand{\shortauthors}{Kerkouri et al.}

\begin{abstract}
Scanpath similarity metrics are central to eye-movement research, yet existing methods predominantly evaluate spatial and temporal alignment while neglecting semantic equivalence between attended image regions. We present a semantic scanpath similarity framework that integrates vision–language models (VLMs) into eye-tracking analysis. Each fixation is encoded under controlled visual context (patch-based and marker-based strategies) and transformed into concise textual descriptions, which are aggregated into scanpath-level representations. Semantic similarity is then computed using embedding-based and lexical NLP metrics and compared against established spatial measures, including MultiMatch and DTW. Experiments on free-viewing eye-tracking data demonstrate that semantic similarity captures partially independent variance from geometric alignment, revealing cases of high content agreement despite spatial divergence. We further analyze the impact of contextual encoding on description fidelity and metric stability. Our findings suggest that multimodal foundation models enable interpretable, content-aware extensions of classical scanpath analysis, providing a complementary dimension for gaze research within the ETRA community. 
\end{abstract}

\ccsdesc[500]{Human-centered computing~Empirical studies in HCI}
\ccsdesc[500]{Computing methodologies~Information extraction}
\ccsdesc[500]{Computing methodologies~Natural language generation}
\ccsdesc[500]{Computing methodologies~Computer vision problems}

\keywords{scanpath similarity, eye tracking, semantic gaze analysis, vision-language models, multimodal AI, large language models, semantic similarity metrics}


\maketitle

\section{Introduction}

Eye tracking captures a high-resolution record of where and when people look, but interpreting \emph{what} they see remains a challenge \cite{tliba2022self}. Classical scanpath similarity metrics, such as MultiMatch\cite{multimatch}, Dynamic Time Warping (DTW)\cite{DWT}, and ScanMatch\cite{scanmatch}, quantify spatial and temporal alignment but ignore the semantic content of attended regions. Two observers may fixate on conceptually similar objects (e.g., faces, text, vehicles) located in different image areas, yielding low spatial similarity despite sharing a common viewing intention. Conversely, similar gaze paths may land on different objects, leading to high geometric similarity but divergent semantic interpretations. This geometric bias limits the utility of scanpath analysis in applications that require understanding of content, such as expertise modeling \cite{tliba2022self} \cite{kerkouri2022, kerkouri2021salypath, kerkouri2022salypath360, kerkouri2026spgen}, user intent inference\cite{jiang2023ueyes}, and adaptive human–AI interaction\cite{mohamed2024review}.

Recent advances in vision-language models (VLMs) \cite{VLMs} offer a new lens: they can translate visual regions into rich natural language descriptions\cite{VLMs}, capturing objects\cite{Feng2025VisionLanguageMF}, attributes, and relationships. By converting each fixation into a short textual description, we can represent a scanpath as a sequence of semantic snapshots. Aggregating these into a coherent summary enables the use of established NLP similarity metrics (e.g., BERTScore \cite{zhang2019bertscore}, ROUGE\cite{lin2004rouge}, BLEU\cite{papineni2002bleu}, BM25\cite{robertson2025bm25}) to compare scanpaths at the level of meaning, not just coordinates. This approach aligns with the goals of generative AI and multimodal systems, where gaze can serve as a semantic signal for personalization, content generation \cite{wang2025target}, and interactive applications\cite{buyukakgul2025vision}.
In this paper, we introduce a first-step framework for semantic scanpath similarity \footnote{The code for this framework will be available at: \href{https://github.com/kmamine/scanpath-semantic-similarity}{https://github.com/kmamine/scanpath-semantic-similarity}.} that integrates VLMs into eye-tracking analysis. Rather than proposing a fully validated new metric, our goal is to explore the feasibility of this direction and identify key design choices. We systematically evaluate two strategies for encoding fixations—local patch extraction and full-image marker annotation—and study how the amount of visual context affects description quality and metric behavior. Using free-viewing eye-tracking data, we compute both semantic similarities (via NLP metrics) and classical spatial similarities (e.g., MultiMatch, DTW) for thousands of scanpath pairs. We then analyze the relationship between these two families of metrics to answer three research questions:

\begin{enumerate}
    \item To what extent does semantic similarity capture variance independent of geometric alignment?
    \item How does the choice of visual context (patch size, marker) influence description fidelity and the resulting semantic similarity?
    \item Can semantic similarity reveal interpretable cases where content agreement diverges from spatial similarity, and what do these cases tell us about gaze behavior?
\end{enumerate}

Our preliminary results show that semantic similarity is partially independent and weakly correlated to spatial metrics, suggesting a new dimension of gaze comparison that may be complementary to geometry. We also find that larger patch sizes improve object recognition but that marker-based encoding may introduce global context confounds. These findings provide initial practical guidance for researchers aiming to incorporate semantic analysis into eye-tracking studies. By re-framing gaze as a semantic modality, this work takes a first step toward the growing intersection of eye tracking and generative AI, opening possibilities for gaze-informed content creation, adaptive interfaces, and human-AI collaboration.

The remainder of this paper is organized as follows. Section \ref{sec:related_works} reviews related work on scanpath metrics and semantic gaze analysis. Section \ref{sec:method} details our method, including fixation encoding, scanpath summarization, and similarity computation. Section \ref{sec:experiments} describes the experimental setup and conditions. Section \ref{sec:results} presents results on description quality, correlation, and divergence, and  discusses implications, limitations, and future directions. Section \ref{sec:conclusion} concludes this paper.

\section{Related Work}
\label{sec:related_works}

\subsection{Spatial and Temporal Scanpath Similarity Metrics}

Quantifying the similarity between two scanpaths is a fundamental task in eye-movement research. Over the past two decades, numerous metrics have been proposed, each capturing different aspects of gaze behavior.

\textbf{Geometric and string-based approaches} treat scanpaths as sequences of fixation points or discretized regions. \emph{ScanMatch}~\cite{scanmatch} divides the stimulus into a grid, assigns each fixation a letter based on its cell, and performs sequence alignment (Needleman–Wunsch) to obtain a similarity score. \emph{MultiMatch}~\cite{multimatch} compares scanpaths along multiple dimensions: shape, direction, length, duration, position, and saccade amplitude, using vector-based matching. \emph{Levenshtein distance}~\cite{levenshtein1966} has also been applied to edit-distance between discretized scanpaths.

\textbf{Time-series alignment methods} accommodate temporal variations. \emph{Dynamic Time Warping (DTW)}~\cite{DWT} aligns two sequences non-linearly, making it robust to differences in fixation duration and scanning speed. \emph{Time-Delay Embedding (TDE)}~\cite{TDE} reconstructs the dynamical system underlying the scanpath and compares trajectories in phase space.

\textbf{Set-based measures} ignore temporal order. \emph{Hausdorff distance} computes the maximum minimal distance between two sets of fixation points, providing a pure spatial dissimilarity measure. \emph{Density-based overlap} (e.g., Kullback–Leibler divergence of fixation maps) is also common in saliency evaluation~\cite{tliba2022satsal, Wong2025Shift, Bruno2023CVD, kerkouri2024AVAtt}.

While these metrics are invaluable for many applications, they all share a fundamental limitation \cite{scanpathmetricreview}: they treat fixations as points in a two-dimensional coordinate space, disregarding the semantic content of the attended regions. Two observers may look at entirely different objects yet receive high spatial similarity if their fixations happen to land near each other; conversely, semantically equivalent objects in different locations yield low spatial similarity. This semantic blindness restricts the interpretability of scanpath comparisons in real-world scenes where meaning matters.

\subsection{Semantic Scanpath Analysis and Vision–Language Models}

Efforts to incorporate semantic information into eye-movement analysis have traditionally relied on manually defined areas of interest (AOIs). Researchers label image regions with categories (e.g., faces, text, objects) and then analyze fixation sequences as transitions between AOIs~\cite{huang2024short, raschke2014visual}. This approach provides semantic insight but is labor-intensive and limited to predefined categories.

With the advent of deep learning, object detectors (e.g., YOLO\cite{ali2024yolo}, Faster R-CNN\cite{fasterrcnn}) have been used to automatically label fixated objects. These methods can identify a wide range of categories, but they remain constrained by the detector's training set and output only class labels, not rich descriptions. Moreover, they treat each fixation independently, losing the narrative flow of the scanpath.

\textbf{Vision–Language Models (VLMs)} have recently emerged as a powerful tool for grounding language in vision. Models such as CLIP~\cite{radford2021learning}, LLaVA~\cite{liu2023llava}, and Qwen-VL~\cite{Qwen-VL} can generate free-form descriptions of image regions, recognize objects, and reason about relationships. They offer the flexibility to describe \emph{any} visual content in natural language, going beyond fixed categories. This capability opens new possibilities for semantic gaze analysis.

A few recent works have begun to leverage VLMs for eye-tracking. For instance,~\cite{mondal2025gaze} used CLIP to compute similarity between fixated patches and text prompts, enabling gaze-based image retrieval. However, these efforts focus on individual fixations or aggregated statistics; they do not convert an entire scanpath into a structured narrative suitable for pairwise similarity computation.

To our knowledge, no prior work has systematically transformed scanpaths into natural language descriptions and used NLP metrics (e.g., BERTScore\cite{zhang2019bertscore}, ROUGE \cite{lin2004rouge}) to compute semantic similarity between gaze sequences. Our framework fills this gap by providing a principled method to encode each fixation with controlled visual context, aggregate them into a coherent summary, and compare scanpaths at the level of meaning. This approach not only complements classical spatial metrics but also aligns eye tracking with the broader trend of multimodal generative AI, where language serves as a unifying representation.

\section{Method}
\label{sec:method}

\subsection{Overview}
\label{sec:method-overview}

\begin{figure}[t]
  \centering
  \includegraphics[width=0.7\columnwidth]{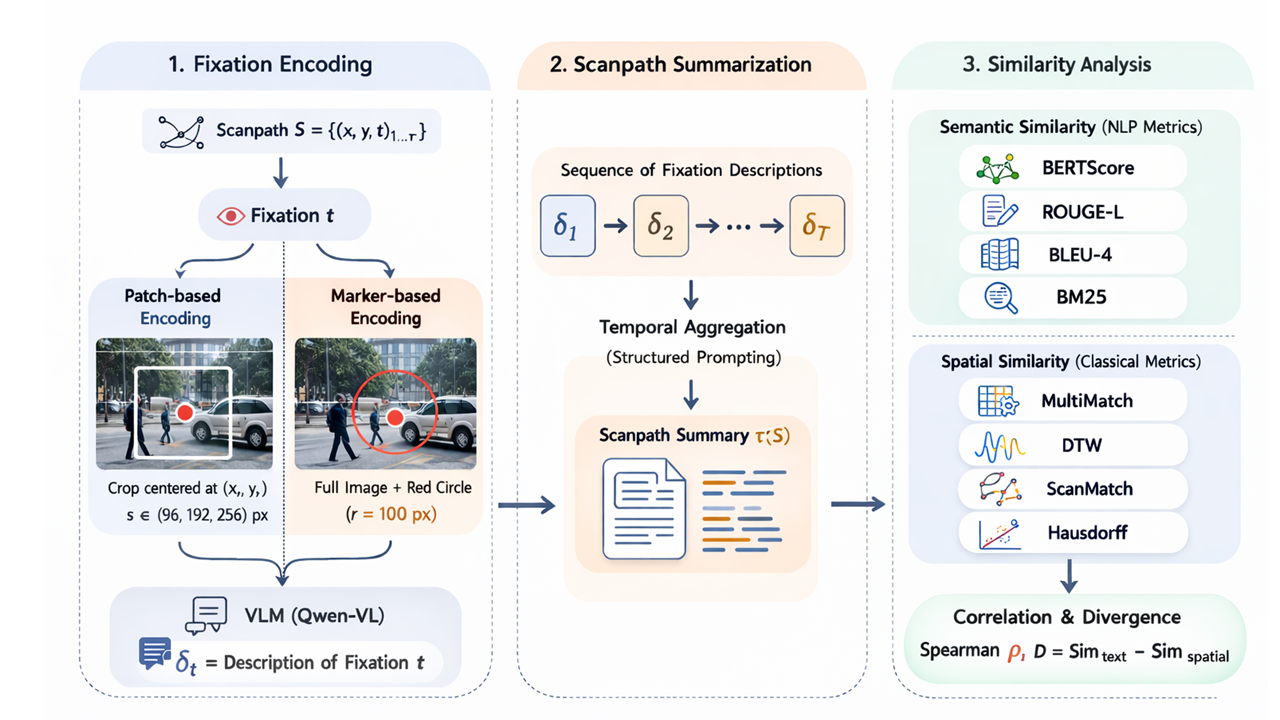}
  \caption{Pipeline overview: fixations are encoded via patch extraction (left) or marker annotation (right), described by a VLM, aggregated into scanpath summaries, then compared via NLP and spatial metrics.}
  \label{fig:pipeline}
  \vspace{-2mm}
\end{figure}

We propose a semantic scanpath similarity pipeline that complements traditional spatial/temporal comparison by explicitly modeling \emph{what} observers looked at. The pipeline (Figure~\ref{fig:pipeline}) converts each fixation into a natural-language description using a Vision-Language Model (VLM), aggregates these into a scanpath-level summary, and computes text similarity between summaries. In parallel, we compute standard spatial metrics from the raw fixation sequences to analyze how semantic and spatial similarity relate.

For a stimulus image $I$ and scanpath $S=\{(x_t,y_t,d_t)\}_{t=1}^{T}$, the pipeline outputs: (1) a scanpath summary $\tau(S)$ describing attended content, and (2) similarity scores $m_{\text{text}}(\tau(S_i),\tau(S_j))$ and $m_{\text{spatial}}(S_i,S_j)$ for scanpath pairs.

The pipeline operates in three stages. \textbf{Stage 1: Fixation-to-text generation.} For each fixation, we generate a description $\delta_t$ using either \emph{patch-based encoding} (cropping a region around the fixation) or \emph{marker-based encoding} (overlaying a circular marker on the full image). \textbf{Stage 2: Scanpath summarization.} The sequence $(\delta_1,\ldots,\delta_T)$ is summarized into a coherent paragraph $\tau(S)$. \textbf{Stage 3: Metric computation and analysis.} We compute semantic similarity via NLP metrics applied to summaries, and spatial similarity via classical scanpath metrics applied to fixation sequences. We then analyze their relationship through correlation and divergence analysis.

This design decouples attended content from geometric/temporal properties, yielding an interpretable semantic representation while retaining compatibility with classical metrics.

\subsection{Fixation-to-Text Generation}
\label{sec:fix-to-text}

We convert each fixation $(x_t,y_t)$ into a short description $\delta_t$ using a VLM under two visual-context encodings. Fixations are first converted to pixel coordinates: $x^{px}_t = \lfloor x_t \cdot W \rfloor$, $y^{px}_t = \lfloor y_t \cdot H \rfloor$.
For each fixation, we extract a square patch $P_t \in \mathbb{R}^{s\times s \times 3}$ centered at $(x^{px}_t,y^{px}_t)$, with $s\in\{96,192,256\}$. Patches extending beyond image boundaries are clamped. Multiple patch sizes test the context/quality trade-off: smaller patches approximate foveal input but may lack object context; larger patches include some peripheral content. We query the VLM with: ``Describe what you see in this image patch in 1-2 sentences. Focus on any objects, faces, text, or salient visual content. If the patch appears blurry or shows only texture/background, describe the dominant colour, texture, or any partial object visible.'' The output is stored as $\delta_t$, constrained to 1--2 sentences to limit verbosity and ensure comparability.

Alternatively, we provide the full image $I$ with a fixation marker: a red circle (radius 100 px, 3 px outline) and center dot (5 px) at $(x^{px}_t,y^{px}_t)$. The marker guides attention without occluding content. We prompt: ``You are analyzing where a viewer looked at an image. The red circle marks the region they fixated on (the circle center is the exact gaze point). Describe what is inside the circled region in 1-2 sentences. Focus on objects or elements within the circle, the visual content at the fixation location, and how this region relates to the broader image context. Be specific about what the viewer was looking at in that circled area.''.  Both encodings map fixations to text but differ in context: patch extraction isolates local evidence; marker annotation provides global context with an explicit pointer.

To compare scanpaths via NLP metrics, we aggregate the temporal sequence $\{\delta_t\}_{t=1}^{T}$ into a single paragraph $\tau(S)$. We format the ordered descriptions as $\texttt{\{fixation\_list\}} = [\delta_1; \delta_2; \ldots; \delta_T]$ and prompt the VLM: ``You are analysing where a human viewer looked at an image. Below are sequential descriptions of the image regions they fixated on (in temporal order): \texttt{\{fixation\_list\}}. Given the full image provided and these fixation descriptions, write a single coherent paragraph summarizing what this viewer attended to and what cognitive strategy they might have used.''.  The output $\tau(S)$ serves as an interpretable semantic representation enabling direct scanpath comparison via text similarity metrics.

\subsection{Metrics and Analyses}
\label{sec:metrics-analyses}

For each within-image scanpath pair $(S_i,S_j)$, we compute semantic similarity from summaries and spatial similarity from fixation sequences, then analyze their relationship.
We compute $\mathrm{Sim}_{text}(S_i,S_j) = m(\tau(S_i),\tau(S_j))$ using four metrics: \textbf{BERTScore} (token-level similarity with contextual embeddings, \texttt{bert-base-uncased}) captures semantic equivalence beyond lexical overlap; \textbf{ROUGE-L} measures longest-common-subsequence overlap; \textbf{BLEU-4} measures $n$-gram precision up to 4-grams (with smoothing); and \textbf{BM25} computes term-frequency–inverse-document-frequency (TF–IDF) weighted lexical relevance, providing a probabilistic retrieval-based measure of content overlap.

We compute $\mathrm{Sim}_{spatial}(S_i,S_j) = g(S_i,S_j)$ using classical metrics spanning sequence alignment, time-series alignment, and set-based distances: \textbf{ScanMatch} (grid-based Needleman--Wunsch), \textbf{Dynamic Time Warping (DTW)} (non-linear alignment), \textbf{MultiMatch} (multi-dimensional feature comparison), \textbf{Hausdorff distance} (spatial proximity), \textbf{Time-Delay Embedding (TDE)} (temporal structure), and \textbf{Levenshtein edit distance} over discretized sequences.

We quantify the semantic-spatial relationship via \textbf{Spearman rank correlation} across scanpath pairs, capturing monotonic relationships without linearity assumptions.
To surface disagreement cases, we define divergence $D(S_i,S_j) = \mathrm{Sim}_{text}(S_i,S_j) - \mathrm{Sim}_{spatial}(S_i,S_j)$. Positive $D$ indicates higher semantic than spatial similarity (similar content at different locations); negative $D$ indicates higher spatial than semantic similarity (similar geometry over different content).
We also track description quality via diagnostics: frequency of blur-related tokens and qualitative inspection of sample descriptions, as semantic similarity depends on description fidelity.

\vspace{-3mm}
\section{Experiments}
\label{sec:experiments}

\subsection{Dataset and Experimental Setup}
\label{sec:dataset-setup}
We evaluate on the COCOFreeView\cite{chen2022characterizing} \cite{yang2023predicting} dataset, which provides free-viewing eye-tracking data over MS-COCO  \cite{lin2014microsoft} images. To enable exhaustive comparisons across conditions, we use a fixed validation subset of 100 images with 5 scanpaths each (500 total scanpaths), randomly sampled from the original validation dataset. For each image, we compute semantic and spatial similarity for all within-image pairs: $\binom{5}{2}=10$ pairs per image, yielding 1000 comparisons per condition. Within-image comparison ensures both scanpaths refer to the same visual content, avoiding cross-image semantic confounds. Stimuli are presented at 1680$\times$1050 resolution under 5-second free viewing.

\subsection{Experimental Conditions}
\label{sec:conditions}

We evaluate four semantic-description conditions varying visual context during fixation description: three patch-based settings (96$\times$96, 192$\times$192, and 256$\times$256 pixel crops) and one marker-based setting (full image with a 100px radius red circle and center dot at the fixation point).

\subsection{Implementation and Procedure}
\label{sec:implementation}

We use \texttt{Qwen/Qwen3-VL-8B-Instruct}\cite{Qwen3-VL} for all generations, running inference with vLLM \cite{kwon2023efficient} on an RTX4000. Fixation descriptions use temperature 0.2 (reducing stochasticity); scanpath summaries use 0.3 (improving fluency). All four conditions run independently.

For each condition and scanpath $S$ on image $I$, we execute:
\setlist{nosep, leftmargin=*}
\begin{enumerate}
  \item \textbf{Fixation encoding:} For each fixation, construct VLM input via (a) centered crop (patch) or (b) full image with marker.
  \item \textbf{Fixation description:} Generate $\delta_t$ (1-2 sentences) of the fixation region.
  \item \textbf{Scanpath summary:} Aggregate $\{\delta_t\}_{t=1}^{T}$ into paragraph $\tau(S)$ (Section~\ref{sec:fix-to-text}).
  \item \textbf{Semantic similarity:} For each within-image pair, compute NLP metrics between summaries (Section~\ref{sec:metrics-analyses}).
  \item \textbf{Spatial similarity:} Compute classical scanpath metrics from fixation sequences (Section~\ref{sec:metrics-analyses}).
  \item \textbf{Analysis-ready outputs:} Store per-pair scores for correlation/divergence analyses.
\end{enumerate}

\begin{figure*}
    \centering
    \includegraphics[width=0.6\linewidth]{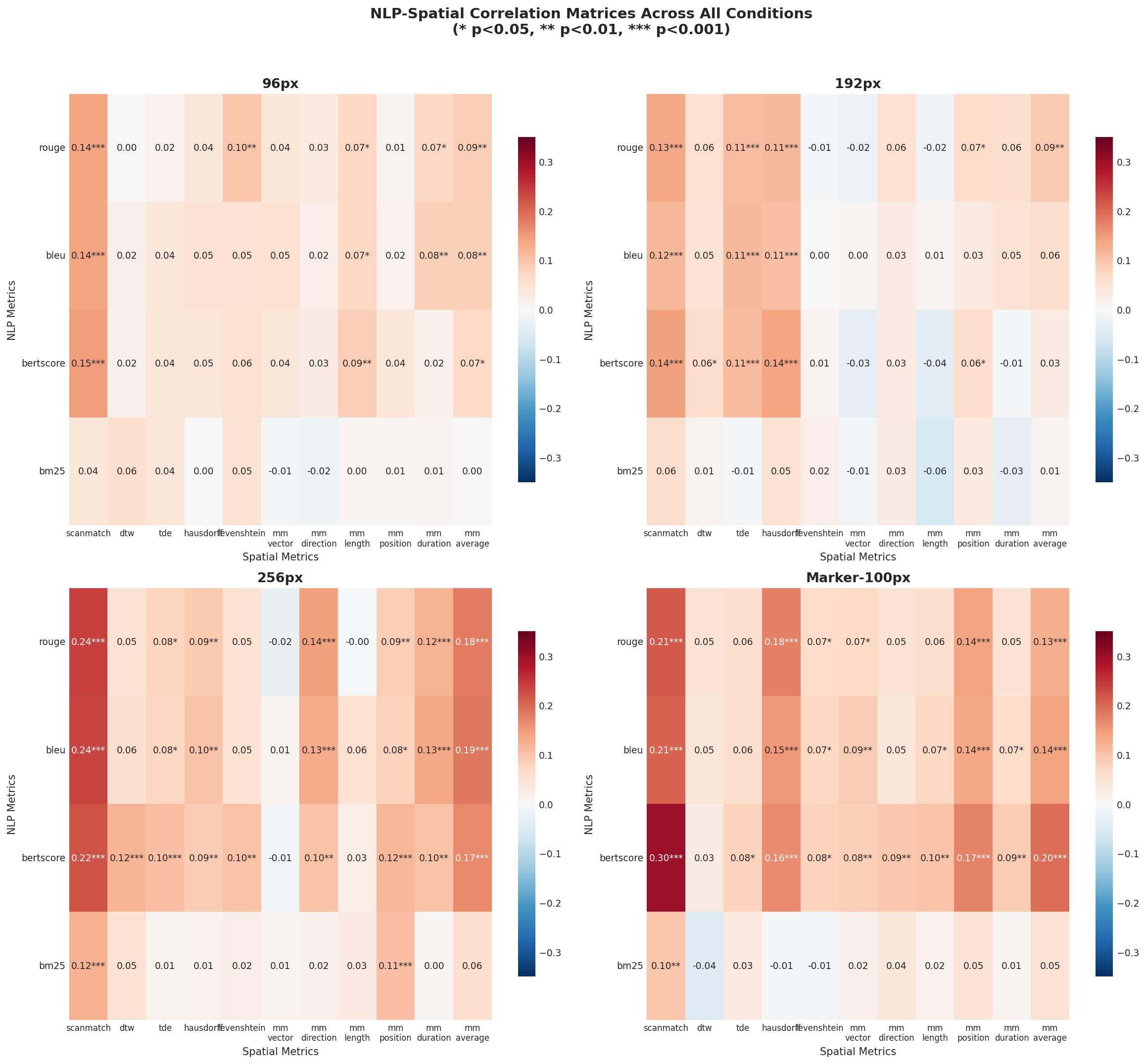}
    \caption{Correlation comparisions between our NLP based metrics framework and spatial/temporal metrics.}
\label{fig:correlation_matrices}
\vspace{-3mm}
\end{figure*}

\section{Results and Discussion}
\label{sec:results}
All analyses are based on 1000 within-image scanpath pairs across four encoding conditions (96px, 192px, 256px, Marker). Figure~\ref{fig:correlation_matrices} presents the full Spearman correlation matrices between semantic similarity metrics (BERTScore, ROUGE-L, BLEU-4, and BM25) and spatial similarity metrics (MultiMatch, DTW, ScanMatch, and a few other metrics) for each condition.
This figure constitutes the central empirical result of the paper.

\subsection{RQ1: Are Semantic and Spatial Similarity Redundant?}

If semantic similarity were merely a reformulation of geometric alignment, correlations between semantic and spatial metrics would approach 1.0 across all conditions.
\textbf{Figure~\ref{fig:correlation_matrices} shows that this is not the case.}
Across patch-based conditions, correlations between BERTScore and spatial metrics fall in the low-moderate range (approximately 0.1–0.3), while lexical metrics (ROUGE-L, BLEU-4) show weaker associations. Importantly:

\begin{itemize}
    \item Correlations are consistently positive, indicating partial coupling.
    \item Correlations remain substantially below 1.0, indicating non-redundancy.
    \item The pattern is stable across all spatial metric families, with ''\textit{ScanMatch}'' presenting the highest correlation values with NLP metrics.
\end{itemize}

This structured moderate correlation confirms that semantic similarity captures substantial variance unexplained by geometric alignment alone. 
Thus, semantic similarity forms a complementary axis of scanpath comparison rather than a surrogate for spatial similarity.

\subsection{RQ2: Effect of Visual Context on Semantic Stability}

Figure~\ref{fig:correlation_matrices} also reveals systematic differences across encoding conditions.

\textbf{Small patches (96px).} Correlations are substantially lower and less stable, particularly for lexical metrics. This reflects reduced object fidelity in fixation descriptions, where limited context produces texture-level or ambiguous language.
\textbf{Intermediate patches (192px).}  Correlations increase and become more consistent, suggesting improved object grounding. \textbf{Large patches (256px).}  Correlations stabilize in the moderate range. This condition yields the clearest and most coherent semantic structure across metrics, indicating reliable object-level encoding without excessive context leakage.

These trends confirm that semantic similarity depends critically on sufficient visual context. Too little context reduces semantic reliability; sufficient local context (\~2\% of the image area) produces stable, interpretable similarity patterns.

\subsection{RQ3: Marker-Based Context Leakage}

The marker condition exhibits a distinct pattern.Compared to the 256px patch condition, marker-based encoding generally produces:

\begin{itemize}
    \item Slightly higher semantic–spatial correlations,
    \item Stronger alignment between BERTScore and spatial metrics,
    \item Reduced spread across semantic metrics.
\end{itemize}

Because the marker condition provides the full image, the VLM can leverage global scene cues when describing fixations. This reduces independence between semantic and spatial similarity by implicitly encoding spatial structure through shared scene context.
The correlation increase in the marker condition therefore suggests semantic leakage: semantic similarity becomes partially inflated by scene-level information beyond the fixation region.

Patch-based encoding, particularly at 256px, better isolates fixation-centered semantics while maintaining sufficient object recognition.

\subsection{Metric-Specific Observations}
The matrices reveal consistent differences across semantic metrics: 
\textbf{BERTScore.}  
Shows the strongest and most stable correlations with spatial metrics. As an embedding-based measure, it captures semantic equivalence beyond lexical overlap while preserving grounding to visual content.

\textbf{ROUGE-L, BLEU-4,and BM25}  
Exhibit weaker and more variable correlations, reflecting sensitivity to surface-form similarity rather than deeper semantic alignment.
This pattern supports the use of embedding-based similarity as the primary semantic metric, with lexical metrics serving as complementary diagnostics.

\subsection{Implications and Limitations}

The correlation structure in Figure~\ref{fig:correlation_matrices} indicates that semantic similarity is neither redundant with nor independent from spatial similarity, but forms a complementary dimension of scanpath comparison. Moderate and stable correlations across patch-based conditions show that semantic representations preserve meaningful grounding in spatial structure while capturing additional content-level variance. The influence of visual context further demonstrates that semantic stability depends on fixation-centered object fidelity, whereas full-image marker encoding reduces independence by introducing global scene cues. These findings position semantic scanpath similarity as a principled extension of geometric metrics, enabling content-aware gaze analysis and facilitating integration of eye-tracking data into multimodal foundation model pipelines.

Several limitations qualify these conclusions. Semantic representations depend on the selected vision–language model and prompting configuration, and no human judgments were collected to directly validate perceived content similarity. The analysis is restricted to within-image free-viewing data, and temporal dynamics are summarized rather than explicitly modeled as structured sequences. Future work should evaluate robustness across multiple VLMs, incorporate human similarity ratings, extend to task-driven and cross-image settings, and develop temporally explicit semantic modeling of scanpaths.



\section{Conclusion}
\label{sec:conclusion}
We introduced a generative AI framework for semantic scanpath similarity that transforms gaze sequences into structured natural language representations using vision–language models. By encoding fixation-centered visual context and comparing scanpaths through embedding-based text similarity, we demonstrated that semantic similarity forms a complementary dimension to classical spatial metrics. The correlation analysis shows moderate but non-redundant alignment between semantic and geometric similarity, while context manipulations reveal how fixation-centered encoding preserves independence and avoids global scene leakage. Together, these results establish that content-level agreement in visual attention can be quantified beyond coordinate overlap.

This work positions gaze as a first-class semantic modality within multimodal AI systems. Translating scanpaths into language enables direct integration with foundation models, interpretable gaze-aware modeling, and new forms of human–AI interaction grounded in attended content rather than location alone. Future research should validate semantic similarity against human judgments, extend the framework to task-driven and cross-image settings, and explore temporally structured semantic modeling, ultimately advancing the convergence of eye tracking and generative multimodal intelligence.

\begin{acks}
  This work has been partially supported by the NIH grants: R01-HL171376 and U01-CA268808.  As well a compute resources from F-Initiatives. 
\end{acks}

\bibliographystyle{ACM-Reference-Format}
\bibliography{sample-base}

@inproceedings{multimatch,
author = {Jarodzka, Halszka and Holmqvist, Kenneth and Nystr\"{o}m, Marcus},
title = {A vector-based, multidimensional scanpath similarity measure},
year = {2010},
isbn = {9781605589947},
publisher = {Association for Computing Machinery},
address = {New York, NY, USA},
url = {https://doi.org/10.1145/1743666.1743718},
doi = {10.1145/1743666.1743718},
booktitle = {Proceedings of the 2010 Symposium on Eye-Tracking Research \& Applications},
pages = {211–218},
numpages = {8},
location = {Austin, Texas},
series = {ETRA '10}
}

@inproceedings{DWT,
author = {Berndt, Donald J. and Clifford, James},
title = {Using dynamic time warping to find patterns in time series},
year = {1994},
publisher = {AAAI Press},
booktitle = {Proceedings of the 3rd International Conference on Knowledge Discovery and Data Mining},
pages = {359–370},
numpages = {12},
location = {Seattle, WA},
series = {AAAIWS'94}
}

@article{scanmatch,
  title={ScanMatch: A novel method for comparing fixation sequences},
  author={Cristino, Filipe and Math{\^o}t, Sebastiaan and Theeuwes, Jan and Gilchrist, Iain D},
  journal={Behavior research methods},
  volume={42},
  number={3},
  pages={692--700},
  year={2010},
  publisher={Springer}
}

@inproceedings{tliba2022self,
  title={Self supervised scanpath prediction framework for painting images},
  author={Tliba, Marouane and Kerkouri, Mohamed Amine and Chetouani, Aladine and Bruno, Alessandro},
  booktitle={Proceedings of the IEEE/CVF Conference on Computer Vision and Pattern Recognition},
  pages={1539--1548},
  year={2022}
}

@inproceedings{kerkouri2024AVAtt,
author = {Kerkouri, Mohamed Amine and Tliba, Marouane and Chetouani, Aladine and Bruno, Alessandro},
title = {AVAtt: Art Visual Attention dataset for diverse painting styles},
year = {2024},
isbn = {9798400706073},
publisher = {Association for Computing Machinery},
address = {New York, NY, USA},
url = {https://doi.org/10.1145/3649902.3655656},
doi = {10.1145/3649902.3655656},
booktitle = {Proceedings of the 2024 Symposium on Eye Tracking Research and Applications},
articleno = {34},
numpages = {3},
location = {Glasgow, United Kingdom},
series = {ETRA '24}
}

@inproceedings{ Bruno2023CVD,
author = {Bruno, Alessandro and Tliba, Marouane and Kerkouri, Mohamed Amine and Chetouani, Aladine and Giunta, Carlo Calogero and \c{C}\"{o}ltekin, Arzu},
title = {Detecting colour vision deficiencies via Webcam-based Eye-tracking: A case study},
year = {2023},
isbn = {9798400701504},
publisher = {Association for Computing Machinery},
address = {New York, NY, USA},
url = {https://doi.org/10.1145/3588015.3590133},
doi = {10.1145/3588015.3590133},
booktitle = {Proceedings of the 2023 Symposium on Eye Tracking Research and Applications},
articleno = {39},
numpages = {2},
location = {Tubingen, Germany},
series = {ETRA '23}
}

@inproceedings{Wong2025Shift,
author = {Wong, David C and Wang, Bin and Durak, Gorkem and Tliba, Marouane and Kerkouri, Mohamed Amine and Chetouani, Aladine and Cetin, Ahmet Enis and Topel, Cagdas and Gennaro, Nicolo and Vendrami, Camila and Agirlar Trabzonlu, Tugce and Rahsepar, Amir Ali and Perronne, Laetitia and Antalek, Matthew and Ozturk, Onural and Okur, Gokcan and Gordon, Andrew C. and Pyrros, Ayis and Miller, Frank H and Borhani, Amir A and Savas, Hatice and Hart, Eric M. and Krupinski, Elizabeth A and Bagci, Ulas},
title = {Shifts in Doctors' Eye Movements Between Real and AI-Generated Medical Images},
year = {2025},
isbn = {9798400714870},
publisher = {Association for Computing Machinery},
address = {New York, NY, USA},
url = {https://doi.org/10.1145/3715669.3726789},
doi = {10.1145/3715669.3726789},
booktitle = {Proceedings of the 2025 Symposium on Eye Tracking Research and Applications},
articleno = {106},
numpages = {7},
location = {
},
series = {ETRA '25}
}

@inproceedings{kerkouri2022,
author = {Kerkouri, Mohamed Amine and Tliba, Marouane and Chetouani, Aladine and Bruno, Alessandro},
title = {A domain adaptive deep learning solution for scanpath prediction of paintings},
year = {2022},
isbn = {9781450397209},
publisher = {Association for Computing Machinery},
address = {New York, NY, USA},
url = {https://doi.org/10.1145/3549555.3549597},
doi = {10.1145/3549555.3549597},
booktitle = {Proceedings of the 19th International Conference on Content-Based Multimedia Indexing},
pages = {57–63},
numpages = {7},
location = {Graz, Austria},
series = {CBMI '22}
}

@article{kerkouri2022salypath360,
author = {Mohamed Amine Kerkouri and Marouane Tliba and Aladine Chetouani and Mohamed Sayeh},
title = {SalyPath360: Saliency and scanpath prediction framework for omnidirectional images},
journal = {Electronic Imaging},
volume = {34},
number = {11},
pages = {168-1--168-1},
doi = {10.2352/EI.2022.34.11.HVEI-168},
url = {https://library.imaging.org/ei/articles/34/11/HVEI-168},
year = {2022},
}

@article{kerkouri2026spgen,
  title={SPGen: Stochastic scanpath generation for paintings using unsupervised domain adaptation},
  author={Kerkouri, Mohamed Amine and Tliba, Marouane and Chetouani, Aladine and Bruno, Alessandro},
  journal={arXiv preprint arXiv:2602.22049},
  year={2026}
}

@INPROCEEDINGS{kerkouri2021salypath,
  author={Kerkouri, Mohamed A. and Tliba, Marouane and Chetouani, Aladine and Harba, Rachid},
  booktitle={2021 IEEE International Conference on Image Processing (ICIP)}, 
  title={Salypath: A Deep-Based Architecture For Visual Attention Prediction}, 
  year={2021},
  volume={},
  number={},
  pages={1464-1468},
  doi={10.1109/ICIP42928.2021.9506295}}

@inproceedings{jiang2023ueyes,
  title={Ueyes: Understanding visual saliency across user interface types},
  author={Jiang, Yue and Leiva, Luis A and Rezazadegan Tavakoli, Hamed and RB Houssel, Paul and Kylm{\"a}l{\"a}, Julia and Oulasvirta, Antti},
  booktitle={Proceedings of the 2023 CHI conference on human factors in computing systems},
  pages={1--21},
  year={2023}
}

@article{mohamed2024review,
  title={A review of machine learning in scanpath analysis for passive gaze-based interaction},
  author={Mohamed Selim, Abdulrahman and Barz, Michael and Bhatti, Omair Shahzad and Alam, Hasan Md Tusfiqur and Sonntag, Daniel},
  journal={Frontiers in Artificial Intelligence},
  volume={7},
  pages={1391745},
  year={2024},
  publisher={Frontiers Media SA}
}

@article{VLMs,
  title={An introduction to vision-language modeling},
  author={Bordes, Florian and Pang, Richard Yuanzhe and Ajay, Anurag and Li, Alexander C and Bardes, Adrien and Petryk, Suzanne and Ma{\~n}as, Oscar and Lin, Zhiqiu and Mahmoud, Anas and Jayaraman, Bargav and others},
  journal={arXiv preprint arXiv:2405.17247},
  year={2024}
}

@article{Feng2025VisionLanguageMF,
  title={Vision-Language Model for Object Detection and Segmentation: A Review and Evaluation},
  author={Yongchao Feng and Yajie Liu and Shuai Yang and Wenrui Cai and Jinqing Zhang and Qiqi Zhan and Ziyue Huang and Hongxi Yan and Qiao Wan and Chenguang Liu and Junzhe Wang and Jiahui Lv and Ziqi Liu and Teng Shi and Qingjie Liu and Yunhong Wang},
  journal={ArXiv},
  year={2025},
  volume={abs/2504.09480},
  url={https://api.semanticscholar.org/CorpusID:277781245}
}

@article{zhang2019bertscore,
  title={Bertscore: Evaluating text generation with bert},
  author={Zhang, Tianyi and Kishore, Varsha and Wu, Felix and Weinberger, Kilian Q and Artzi, Yoav},
  journal={arXiv preprint arXiv:1904.09675},
  year={2019}
}

@inproceedings{lin2004rouge,
  title={Rouge: A package for automatic evaluation of summaries},
  author={Lin, Chin-Yew},
  booktitle={Text summarization branches out},
  pages={74--81},
  year={2004}
}

@inproceedings{papineni2002bleu,
  title={Bleu: a method for automatic evaluation of machine translation},
  author={Papineni, Kishore and Roukos, Salim and Ward, Todd and Zhu, Wei-Jing},
  booktitle={Proceedings of the 40th annual meeting of the Association for Computational Linguistics},
  pages={311--318},
  year={2002}
}

@inproceedings{robertson2025bm25,
  title={BM25 and all that--a look back},
  author={Robertson, Stephen},
  booktitle={Proceedings of the 48th International ACM SIGIR Conference on Research and Development in Information Retrieval},
  pages={5--8},
  year={2025}
}

@inproceedings{wang2025target,
  title={Target scanpath-guided 360-degree image enhancement},
  author={Wang, Yujia and Zhang, Fang-Lue and Dodgson, Neil A},
  booktitle={Proceedings of the AAAI Conference on Artificial Intelligence},
  volume={39},
  number={8},
  pages={8169--8177},
  year={2025}
}

@article{buyukakgul2025vision,
  title={Where Vision Meets Memory: An Eye-Tracking Study of In-App Ads in Mobile Sports Games with Mixed Visual-Quantitative Analytics},
  author={B{\"u}y{\"u}kakg{\"u}l, {\"U}mit Can and Y{\"u}ce, Arif and Kat{\i}rc{\i}, Hakan},
  journal={Journal of Eye Movement Research},
  volume={18},
  number={6},
  pages={74},
  year={2025},
  publisher={MDPI}
}

@inproceedings{levenshtein1966,
  title={Binary codes capable of correcting deletions, insertions, and reversals},
  author={Levenshtein, Vladimir I and others},
  booktitle={Soviet physics doklady},
  volume={10},
  number={8},
  pages={707--710},
  year={1966},
  organization={Soviet Union}
}

@article{TDE,
  title={Machine learning of time series using time-delay embedding and precision annealing},
  author={Ty, Alexander JA and Fang, Zheng and Gonzalez, Rivver A and Rozdeba, Paul J and Abarbanel, Henry DI},
  journal={Neural Computation},
  volume={31},
  number={10},
  pages={2004--2024},
  year={2019},
  publisher={MIT Press One Rogers Street, Cambridge, MA 02142-1209, USA journals-info~…}
}

@article{tliba2022satsal,
  title={Satsal: A multi-level self-attention based architecture for visual saliency prediction},
  author={Tliba, Marouane and Kerkouri, Mohamed A and Ghariba, Bashir and Chetouani, Aladine and {\c{C}}{\"o}ltekin, Arzu and Shehata, Mohamed Sami and Bruno, Alessandro},
  journal={IEEE Access},
  volume={10},
  pages={20701--20713},
  year={2022},
  publisher={IEEE}
}

@article{huang2024short,
  title={Short-time AOIs-based representative scanpath identification and scanpath aggregation},
  author={Huang, He and Doebler, Philipp and Mertins, Barbara},
  journal={Behavior Research Methods},
  volume={56},
  number={6},
  pages={6051--6066},
  year={2024},
  publisher={Springer}
}

@inproceedings{raschke2014visual,
  title={A visual approach for scan path comparison},
  author={Raschke, Michael and Herr, Dominik and Blascheck, Tanja and Ertl, Thomas and Burch, Michael and Willmann, Sven and Schrauf, Michael},
  booktitle={Proceedings of the symposium on eye tracking research and applications},
  pages={135--142},
  year={2014}
}

@article{ali2024yolo,
  title={The YOLO framework: A comprehensive review of evolution, applications, and benchmarks in object detection},
  author={Ali, Momina Liaqat and Zhang, Zhou},
  journal={Computers},
  volume={13},
  number={12},
  pages={336},
  year={2024},
  publisher={MDPI}
}

@inproceedings{fasterrcnn,
author = {Ren, Shaoqing and He, Kaiming and Girshick, Ross and Sun, Jian},
title = {Faster R-CNN: towards real-time object detection with region proposal networks},
year = {2015},
publisher = {MIT Press},
address = {Cambridge, MA, USA},
booktitle = {Proceedings of the 29th International Conference on Neural Information Processing Systems - Volume 1},
pages = {91–99},
numpages = {9},
location = {Montreal, Canada},
series = {NIPS'15}
}

@inproceedings{radford2021learning,
  title={Learning transferable visual models from natural language supervision},
  author={Radford, Alec and Kim, Jong Wook and Hallacy, Chris and Ramesh, Aditya and Goh, Gabriel and Agarwal, Sandhini and Sastry, Girish and Askell, Amanda and Mishkin, Pamela and Clark, Jack and others},
  booktitle={International conference on machine learning},
  pages={8748--8763},
  year={2021},
  organization={PmLR}
}

@misc{liu2023llava,
      title={Visual Instruction Tuning}, 
      author={Liu, Haotian and Li, Chunyuan and Wu, Qingyang and Lee, Yong Jae},
      publisher={NeurIPS},
      year={2023},
}

@article{Qwen3-VL,
      title={Qwen3-VL Technical Report}, 
      author={Shuai Bai and Yuxuan Cai and Ruizhe Chen and Keqin Chen and Xionghui Chen and Zesen Cheng and Lianghao Deng and Wei Ding and Chang Gao and Chunjiang Ge and Wenbin Ge and Zhifang Guo and Qidong Huang and Jie Huang and Fei Huang and Binyuan Hui and Shutong Jiang and Zhaohai Li and Mingsheng Li and Mei Li and Kaixin Li and Zicheng Lin and Junyang Lin and Xuejing Liu and Jiawei Liu and Chenglong Liu and Yang Liu and Dayiheng Liu and Shixuan Liu and Dunjie Lu and Ruilin Luo and Chenxu Lv and Rui Men and Lingchen Meng and Xuancheng Ren and Xingzhang Ren and Sibo Song and Yuchong Sun and Jun Tang and Jianhong Tu and Jianqiang Wan and Peng Wang and Pengfei Wang and Qiuyue Wang and Yuxuan Wang and Tianbao Xie and Yiheng Xu and Haiyang Xu and Jin Xu and Zhibo Yang and Mingkun Yang and Jianxin Yang and An Yang and Bowen Yu and Fei Zhang and Hang Zhang and Xi Zhang and Bo Zheng and Humen Zhong and Jingren Zhou and Fan Zhou and Jing Zhou and Yuanzhi Zhu and Ke Zhu},
	  journal={arXiv preprint arXiv:2511.21631},
      year={2025}
}

@article{Qwen-VL,
  title={Qwen-VL: A Versatile Vision-Language Model for Understanding, Localization, Text Reading, and Beyond},
  author={Bai, Jinze and Bai, Shuai and Yang, Shusheng and Wang, Shijie and Tan, Sinan and Wang, Peng and Lin, Junyang and Zhou, Chang and Zhou, Jingren},
  journal={arXiv preprint arXiv:2308.12966},
  year={2023}
}

@inproceedings{mondal2025gaze,
  title={Gaze-Language Alignment for Zero-Shot Prediction of Visual Search Targets from Human Gaze Scanpaths},
  author={Mondal, Sounak and Sendhilnathan, Naveen and Zhang, Ting and Liu, Yue and Proulx, Michael and Iuzzolino, Michael Louis and Qin, Chuan and Jonker, Tanya R},
  booktitle={Proceedings of the IEEE/CVF International Conference on Computer Vision},
  pages={2738--2749},
  year={2025}
}

@inproceedings{scanpathmetricreview,
author = {Duchowski, Andrew T. and Driver, Jason and Jolaoso, Sheriff and Tan, William and Ramey, Beverly N. and Robbins, Ami},
title = {Scanpath comparison revisited},
year = {2010},
isbn = {9781605589947},
publisher = {Association for Computing Machinery},
address = {New York, NY, USA},
url = {https://doi.org/10.1145/1743666.1743719},
doi = {10.1145/1743666.1743719},
booktitle = {Proceedings of the 2010 Symposium on Eye-Tracking Research \& Applications},
pages = {219–226},
numpages = {8},
location = {Austin, Texas},
series = {ETRA '10}
}

@inproceedings{chen2022characterizing, 

title={Characterizing Target-Absent Human Attention}, 

author={Chen, Yupei and Yang, Zhibo and Chakraborty, Souradeep and Mondal, Sounak and Ahn, Seoyoung and Samaras, Dimitris and Hoai, Minh and Zelinsky, Gregory}, 

booktitle={Proceedings of the IEEE/CVF Conference on Computer Vision and Pattern Recognition Workshops}, 

pages={5031--5040}, 

year={2022} 

}

@article{yang2023predicting,

title={Predicting Human Attention using Computational Attention},

author={Yang, Zhibo and Mondal, Sounak and Ahn, Seoyoung and Zelinsky, Gregory and Hoai, Minh and Samaras, Dimitris},

journal={arXiv preprint arXiv:2303.09383},

year={2023}

}

@inproceedings{lin2014microsoft,
  title={Microsoft coco: Common objects in context},
  author={Lin, Tsung-Yi and Maire, Michael and Belongie, Serge and Hays, James and Perona, Pietro and Ramanan, Deva and Doll{\'a}r, Piotr and Zitnick, C Lawrence},
  booktitle={European conference on computer vision},
  pages={740--755},
  year={2014},
  organization={Springer}
}

@inproceedings{kwon2023efficient,
  title={Efficient Memory Management for Large Language Model Serving with PagedAttention},
  author={Woosuk Kwon and Zhuohan Li and Siyuan Zhuang and Ying Sheng and Lianmin Zheng and Cody Hao Yu and Joseph E. Gonzalez and Hao Zhang and Ion Stoica},
  booktitle={Proceedings of the ACM SIGOPS 29th Symposium on Operating Systems Principles},
  year={2023}
}


\end{document}